\documentclass[preprint,5p,times,twocolumn,authoryear]{elsarticle}

\usepackage{amssymb}
\usepackage{amsmath}
\usepackage[colorlinks=true, linkcolor=blue, citecolor=blue]{hyperref}
\usepackage{amsfonts}
\usepackage{algorithmic}
\usepackage{algorithm}
\usepackage{array}
\usepackage{pifont}
\usepackage{textcomp}
\usepackage{stfloats}
\usepackage{url}
\usepackage{verbatim}
\usepackage{graphicx}
\usepackage{cite}
\usepackage{bbm}
\usepackage{booktabs}
\usepackage{multirow}
\usepackage{xcolor}
\usepackage{xcolor}
\usepackage{amssymb}
\usepackage{tablefootnote}
\usepackage{amsmath}
\usepackage{subfigure}
\usepackage{amsfonts}
\usepackage{wrapfig}
\usepackage{bm}
\usepackage{caption}
\usepackage{subcaption}

\begin{document}

\begin{frontmatter}

\title{{ProgD}: Progressive Multi-scale Decoding with Dynamic Graphs for Joint Multi-agent Motion Forecasting}

\author[shailab]{Xing Gao\corref{cor1}}
\author[sjtu]{Zherui Huang}
\author[sjtuee]{Weiyao Lin}
\author[shailab]{Xiao Sun}
\affiliation[shailab]{organization={Shanghai Artificial Intelligence Laboratory},
            city={Shanghai},
            postcode={200232}, 
            country={China}}
\affiliation[sjtu]{organization={Department of Computer Science, Shanghai Jiao Tong University},
            city={Shanghai},
            postcode={200240}, 
            country={China}}
\affiliation[sjtuee]{organization={Department of Electronic Engineering, Shanghai Jiao Tong University},
            city={Shanghai},
            postcode={200240}, 
            country={China}}
\cortext[cor1]{Corresponding author, e-mail: gxyssy@163.com.}

\begin{abstract}
Accurate motion prediction of surrounding agents is crucial for the safe planning of autonomous vehicles. Recent advancements have extended prediction techniques from individual agents to joint predictions of multiple interacting agents, with various strategies to address complex interactions within future motions of agents. However, these methods overlook the evolving nature of these interactions.  To address this limitation, we propose a novel progressive multi-scale decoding strategy, termed ProgD, with the help of dynamic heterogeneous graph-based scenario modeling. In particular, to explicitly and comprehensively capture the evolving social interactions in future scenarios, given their inherent uncertainty, we design a progressive modeling of scenarios with dynamic heterogeneous graphs. With the unfolding of such dynamic heterogeneous graphs, a factorized architecture is designed to process the spatio-temporal dependencies within future scenarios and progressively eliminate uncertainty in future motions of multiple agents. Furthermore, a multi-scale decoding procedure is incorporated to improve on the future scenario modeling and consistent prediction of agents' future motion. The proposed ProgD achieves state-of-the-art performance on the INTERACTION multi-agent prediction benchmark, ranking $1^{st}$, and the Argoverse~2 multi-world forecasting benchmark.
\end{abstract}


\end{frontmatter}

\section{Introduction}
\label{sec1}

Motion prediction is important for self-driving systems to ensure safe and efficient navigation. Of particular interest is joint multi-agent motion prediction, which involves concurrently forecasting the future trajectories of all agents within a scene. This task has gained increasing attention recently, due to its complexity compared to marginal motion prediction, as it requires maintaining consistency and coherence in future motions of interactive agents, reflecting the intricate dynamics of real-world traffic. Without such consistency, the prediction module could produce conflicting trajectories, such as collisions between the predicted motions of agents, which would undermine the reliability of the system and lead to unsafe or infeasible motion plans. 

The challenge is further compounded by several intrinsic factors:  (1) Dynamic and complex future social interactions.  Agent engages in various types of interactions with surrounding vehicles, pedestrians, and roads, and these interactions continually evolve as the agents progress. For instance,  at a crossroad, considering the dynamic interaction near the intersection is crucial to avoid collisions when vehicle A intends to make a left turn while vehicle B plans to go straight. (2) Scalability. With an increasing number of agents, the joint trajectory space expands exponentially \citep{rowe2023fjmp}, resulting in increasingly complex interactions among them.

Existing methods have attempted to address these challenges through various architectures. For example, several approaches based on various graph convolution networks  \citep{mo2020recog,an2022dginet} and Transformer-based  frameworks~\citep{zhou2022hivt,JiaWCLLY23,zhang2024edge} effectively model observed interactions and incorporate encoded social interaction features for joint prediction during the decoding stage.  However, these methods primarily focus on modeling observed interactions while fundamentally ignoring interactions in future states. Advancing beyond this limitation, some works have explored the implicit modeling of future interactions for joint trajectory prediction.  For instance, attention schemes are employed in the decoder of \citep{ngiam2021scene,Jiang_2023_CVPR,shi2023mtr++} to learn the joint characteristics of agents, and latent variable models are used in \citep{casas2020implicit,suo2021trafficsim,GirgisGCWDKHP22} to capture unobserved dynamics. More explicitly, several recent works resort to structured models like graphs to model interactions in future scenarios  \citep{luo2023jfp,rowe2023fjmp}. However, these methods typically assume future interactions as static, thus failing to capture their inherent dynamic evolution throughout the prediction horizon.

\begin{figure}[tp]
\begin{center}
\includegraphics[width=0.45\textwidth]{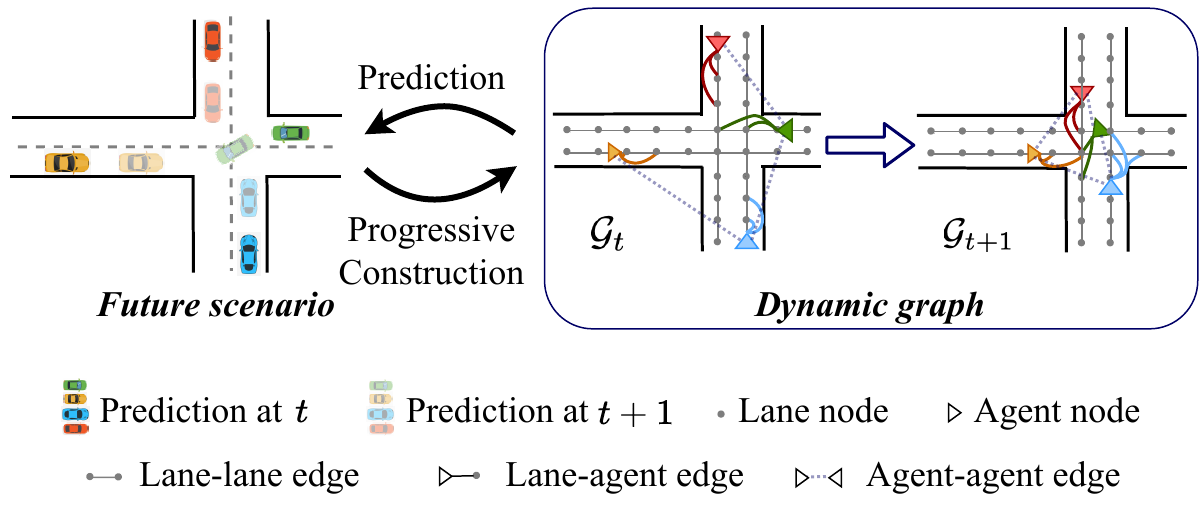}
\end{center}
\caption{High-level illustration of joint motion prediction through progressive modeling of future scenarios with dynamic graphs. The dynamic graph renders a structured representation for predicting the agent states at the next time steps.  Based on these predicted states,  the subsequent snapshot of the dynamic graph is further constructed to capture updated interactions. This progressive prediction process continues until the end of the prediction horizon.} \label{f:0}
\end{figure}

To bridge this gap,  we propose a joint multi-agent motion forecasting method, termed ProgD. Our core insight is that modeling the progressive evolution of future interactions is essential for accurate and consistent joint prediction. In particular,  ProgD exploits dynamic graphs that explicitly evolve over future time steps, continuously capturing the spatio-temporal dependencies between agents and their environment as the scenario unfolds,   accommodating  an arbitrary number of agents, as illustrated in Fig.~\ref{f:0}. This dynamic structure enables ProgD to progressively process encoded scene information and gradually decode multimodal, consistent joint future trajectories in a multi-scale manner, effectively addressing  uncertainty, scalability, and the continuous dynamics of future interactions. 

The benefits of ProgD are summarized as follows.
\begin{itemize}
    \item The proposed innovative progressive modeling strategy of future scenarios  utilizes dynamic heterogeneous graphs to explicitly encode the diverse interactions among agents and their interactions with the road network in future scenarios, accounting for their dynamic changes over time.
    \item  Through unfolding dynamic graphs, ProgD gradually eliminates the inherent uncertainty in future interacting motions of multiple agents. Consequently, in conjunction with multi-scale decoding, it effectively leverages the predicted information to enhance the accuracy and consistency of predictions.
    \item ProgD enables the joint prediction of consistent future motions for an arbitrary number of agents. It achieves state-of-the-art performance on the INTERACTION multi-agent (\textbf{the $\bm{1^{st}}$ ranking)} and Argoverse 2 multi-world forecasting benchmarks. 
\end{itemize}

In more detail, the proposed ProgD seamlessly integrates the dynamic graph-based modeling of future scenarios and motion prediction of interacting agents. In contrast to the modeling of observed scenarios, future scenarios are characterized by uncertainty, making it infeasible to directly construct dynamic graphs. As a result, we adopt a progressive construction approach that dynamically builds the graph in synchronization with the prediction of agent motions. Specifically, the predicted future motions of agents are utilized to incrementally construct the next snapshot of dynamic graphs, effectively encoding the corresponding interactions. Notably, scene elements including agents as well as roads are all taken into consideration, and their diverse interactions are encoded as heterogeneous edges. This information is further processed using tailored heterogeneous graph convolutional networks to predict subsequent motions. Furthermore, we incorporate a multi-scale decoding strategy into the progressive process.  At each step, the representations obtained from the heterogeneous graph convolutional networks are utilized to make a coarse estimation of the last positions for the multiple agents in the next time interval. These coarse predictions guide the updating of the current snapshot of the dynamic graph, adjusting node features and connections to reflect evolving interactions. Building upon the features from the updated dynamic graph, we proceed to perform a fine-grained prediction of the complete states of the agents in the upcoming time interval. By continuously unfolding such graph construction and prediction steps, we gradually eliminate the uncertainty of future scenarios, effectively capturing the complex dynamics and interactions within them. This iterative multi-scale process allows for obtaining accurate and consistent predictions of future motions of the multi-agents.

\begin{figure*}[tp]
\begin{center}
\includegraphics[width=\textwidth]{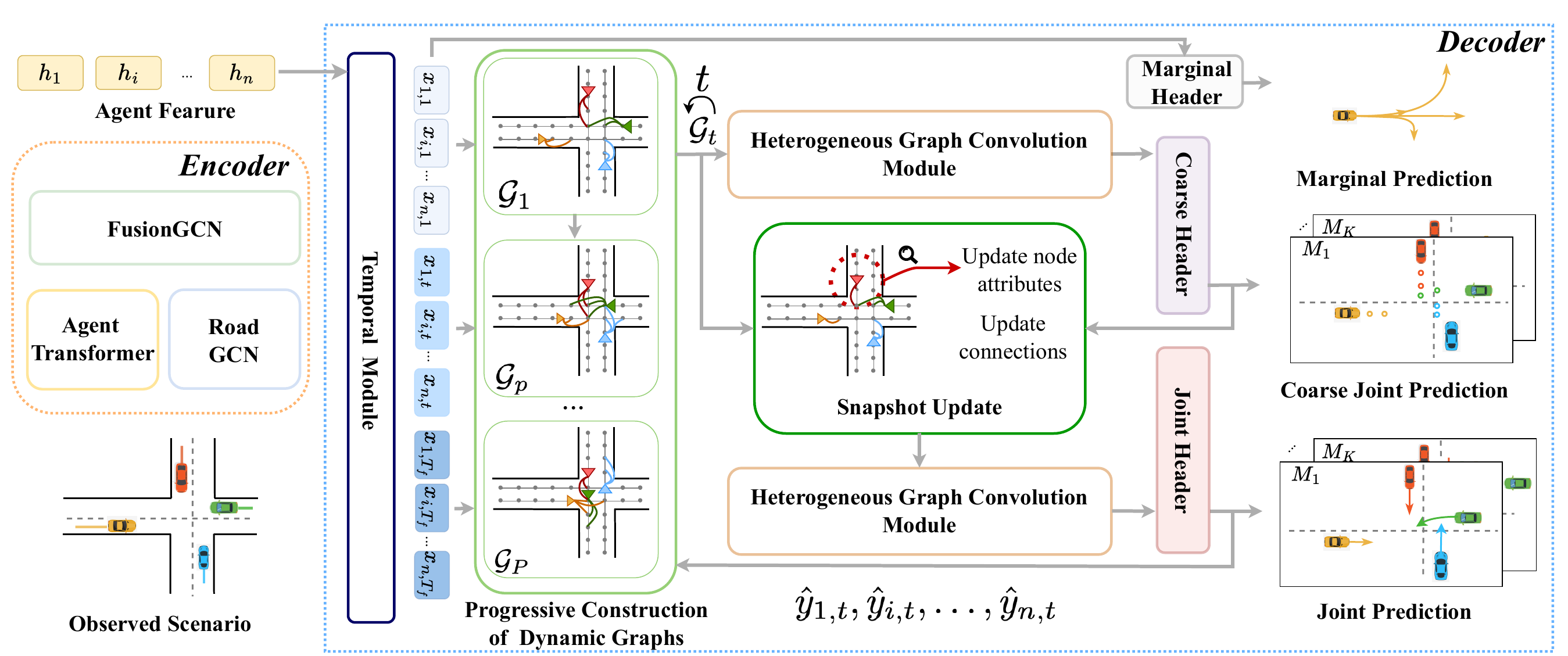}
\end{center}
\caption{Illustration of the encoder-decoder framework of ProgD. The encoder encodes the observed trajectories of agents and road networks in a scenario, and outputs the features of agents $\bm{h_1}, \bm{h_i}, \dots, \bm{h_n}$. The decoder consists of a temporal module, two graph convolution modules for dual-stage dynamic graphs, and several headers responsible for decoding the marginal and joint future states of interacting agents, $\bm{\hat{y}_{1,t}}, \bm{\hat{y}_{i,t}}, \dots, \bm{\hat{y}_{n,t}}$, within a scenario. The temporal module is designed to capture the temporal dependencies in the future motions of each agent, while the two graph convolution modules handle the evolving interactions among multiple agents in their future motions, which are captured with dynamic graphs.} \label{f:1}
\end{figure*}

Effectively addressing the interleaved future spatio-temporal information is a critical component of such a decoding procedure. To overcome this challenge, we introduce an encoder-decoder architecture.
Specially, in the decoder, we introduce a factorization approach to handle future spatio-temporal information, tailored for the progressive decoding strategy. As shown in Fig.~\ref{f:1}, based on the representations of agents' historical states and road network structure from the encoder, we employ a temporal model as the initial step to predict the temporal feature sequences of all agents. This temporal model aims to effectively exploit the encoded temporal dependencies and motion patterns, accurately forecasting the evolution of temporal behaviors.  Furthermore, graph convolutional networks are designed to handle the complex interactions within the previously predicted states of the agents,  encoded in the dynamic graphs, resulting in spatio-temporally consistent multi-agent features for subsequent motion forecasting.

We evaluate the proposed ProgD strategy on two widely-used real-world multi-agent joint motion forecasting benchmarks, Argoverse 2 and INTERACTION. These datasets encompass various complex traffic scenarios that involve multiple interacting agents to jointly predict, and provide a comprehensive set of evaluation metrics to assess the accuracy and consistency of predictions. Various evaluations demonstrate that ProgD outperforms the state-of-the-art methods on both datasets. Particularly, it achieves the top-ranking position on the INTERACTION Multi-Agent Leaderboard.

\section{Related Works}

\textbf{Marginal motion prediction.} In most cases, marginal motion prediction approaches typically employ advanced encoder architectures to process social interactions and extract spatio-temporal features from observed scenarios. These features are subsequently fed into lightweight decoders to predict the future trajectories of individual agents.  Generally, existing methods can be categorized into two paradigms  in terms of scenario representation learning: rasterization-based strategies and vectorized methods.  First, rasterization strategies \citep{chai2019multipath, casas2018intentnet, hong2019rules, bansal2018chauffeurnet,  cui2019multimodal} learn dense representations of scenarios by encoding past agent trajectories and road networks into bird’s-eye-view images. Convolutional neural networks are then employed to extract relevant features from these images. However, rasterization-based methods are inefficient in addressing long-range interactions and capturing the sparse structure of road networks. To address these limitations,  vectorized methods represent observed scenarios using graphs or point sets to model various types of interactions and  learn sparse encodings.  For instance,  graph neural network (GNN)-based methods  have been proposed to represent interaction networks among  agents~\citep{mo2022multi}, road networks~\citep{liang2020learning,gilles2021gohome,zeng2021lanercnn}, or whole observed scenes~\citep{gao2020vectornet,gao2023dynamic}. Alternatively, a series of methods represent historical trajectories and road networks as point sets or polylines, exploiting point cloud-based techniques \citep{ye2021tpcn,ye2022dcms} or Transformer-based architectures \citep{nayakanti2023wayformer,varadarajan2021multipath,FengZLZXZZS23,zhang2024real,AzadaniB25} to capture complex interactions among agents and road lanes within the observed scenarios.

Furthermore, several methods explore various methods to improve the decoding procedure. For instance, DensenTNT \citep{gu2021densetnt} employ optimization techniques to predict trajectories based on a dense goal candidate set. Although marginal prediction approaches can produce plausible predictions for individual agents, they often result in inconsistent trajectories when multiple interactive agents are considered.

\textbf{Joint multi-agent prediction.} Some recent works have extended from marginal prediction to joint prediction of multiple interacting agents by directly incorporating observed social interaction features during the decoding stage.  While sharing this paradigm, various methods differ in their designs for capturing and representing social interactions within observed scenarios. For example, ReCoG2 \citep{mo2020recog} utilizes a heterogeneous graph encoder to encode historical trajectories of agents and map information, and combines interaction features and the target vehicle’s sequential features for a RNN decoder to predict joint trajectories. Similarly, HiVT~\citep{zhou2022hivt}  exploits a hierarchical vector Transformer to enhance local context and global interaction modeling. HDGT \citep{JiaWCLLY23} and EEGT~\citep{zhang2024edge} further improve the modeling of interaction within the observed scenario by introducing heterogeneous graph Transformers and edge-enriched graph Transformers, respectively.   Besides, Forecast-MAE \citep{Cheng_2023_ICCV} and Traj-MAE \citep{ChenWSLHGCH23} design diverse self-supervised masking strategies with Transformer architectures to better capture social interaction information. However, despite their efforts to enhance the modeling and representation of interactions within observed scenarios, these methods primarily focus on the observed domain and generally neglect the modeling of interactions among agents in their future motions.

Then, several works resort to implicitly modeling future interactions, such as by employing agent-wise attention mechanisms in the decoder and utilizing latent variable models. For example, SceneTransformer~\citep{ngiam2021scene}, MotionDiffuser~\citep{Jiang_2023_CVPR}, and QCNet~\citep{zhou2023query} perform joint prediction through alternative attention layers across time and agents in the decoder. In addition, MotionLM~\citep{seff2023motionlm} represents continuous trajectories as sequences of discrete motion tokens, and generates joint trajectories with an autoregressive Transformer decoder by simultaneously sampling tokens for interacting agents while enabling them to attend to each other iteratively. MTR++ \citep{shi2023mtr++} further considers the relative spatial relationship of intention queries among every pair of agents in the attention layers to make scenario-compliant predictions for multiple agents.  Alternatively, ILVM~\citep{casas2020implicit}, TrafficSim~\citep{suo2021trafficsim}, and  AutoBots \citep{GirgisGCWDKHP22} leverage latent variable models to implicitly characterize uncertainties and interactions in future scenarios. In contrast to these methods that implicitly model future interactions, our approach explicitly represents future scenarios using dynamic graphs, which allows for more precise and interpretable modeling of the evolving interactions among agents.

In addition, some works exploit marginal predictions to produce joint forecasting. For instance,  THOMAS \citep{gilles2021gohome}  employs attention schemes to recombine candidate marginal predictions of each agent to obtain consistent joint predictions. Similarly, JFP \citep{luo2023jfp} constructs an interactive graph of agents based on the closeness of the marginal candidates of each pair of agents and exploits message passing to approximate joint probability. Moreover, FFINet \citep{kang2024ffinet} introduces a feedback module to consider the influence of future interactions in initial predictions, and DGFNet~\citep{xin2025multi} considers the prediction difficulty among agents in decoding.

Furthermore, several methods employ conditional models to make joint predictions. For example, M2I \citep{sun2022m2i} first identifies the influencer and reactor of two agents and then predicts the reactor's future motion based on the influencer's marginal forecast. However, it only considers interactions between just two agents. FJMP \citep{rowe2023fjmp} further exploits a directed acyclic graph to model the interactions among an arbitrary number of agents and use graph neural networks to perform marginal and conditional inference. However, these models are still limited to using static graphs to handle future interactions, which continuously evolve as agents navigate, particularly in tasks with long prediction horizons.

Contrary to these methods, the proposed ProgD further considers the evolution of future interactions with a dynamic graph rather than a static graph. Through continuously exploring and capturing the surrounding environment and social interactions, ProgD employs a progressive decoding strategy to gradually unfold future scenarios and effectively handle the dynamic spatio-temporal information of future motions. Furthermore, through multi-scale predictions ranging from coarse-grained to fine-grained levels, ProgD ensures the consistency and accuracy of joint multi-agent motion predictions.

\section{Methodology}
To improve the \textbf{consistency of future motions among interactive agents}, we propose ProgD, a novel framework that explicitly models the evolution of future interactions through \textit{dynamic heterogeneous graphs}. By continuously capturing the spatio-temporal dependencies in future scenarios with dynamic graphs, ProgD flexibly accommodates arbitrary numbers of agents and effectively represents the complex, evolving nature of real-world traffic,  as elaborated in Sec.~3.3.1. This progressive modeling strategy, together with  \textit{multi-scale decoding}, enables accurate and consistent joint trajectory prediction, as detailed in Sec.~3.3.2. 

\subsection{Preliminaries}
We aim to jointly forecast the motion of $n$ agents in the future $T_f$ timesteps in a scenario based on their historical states of past $T_o$ seconds and road networks. For each scenario, we make $K$ joint predictions of the interesting agents in the scenario, with each joint prediction referred to as a modality.~\footnote{In the motion prediction literature, modality often refers to multiple plausible future trajectories that capture the inherent uncertainty of prediction,  in contrast to multimodal learning where it refers to different input data types such as text, image, and audio.} To achieve this, we adopt an encoder-decoder framework. The encoder embeds the historical information of agents and road networks, while the decoder forecasts $K$ joint future motions of the multiple interesting agents in the scenario, as illustrated in Fig.~\ref{f:1}.

\textbf{Notions.} For each scenario, we take the last observation time as $t=0$, and the observation period and prediction horizon correspond to $[-T_{o+1}, 0]$ and $[1, T_f]$, respectively. The observed states and predicted states of an arbitrary agent are denoted by $S_i = [\bm{s_{i,-T_o+1}}, \dots,  \bm{s_{i,-1}}, \bm{s_{i,0}}]$ and $\hat{Y}_i = [\bm{\hat{y}_{i,1}}, \bm{\hat{y}_{i,2}} \dots, \bm{\hat{y}_{i, T_f}}]$,  with $\bm{\cdot_{i,t}}$ indicating the state of agent $i$ at time $t$, $i=1,2,\dots, n$. In this paper, we employ dynamic heterogeneous graphs to model future scenarios. A discrete-time dynamic graph $\mathcal{G}$ typically consists of a collection of discrete snapshots $\mathcal{G}_p= \{\mathcal{V}_p, \mathcal{E}_p\}_{p=1}^{P}$, with $\mathcal{V}_p$ and  $\mathcal{E}_p$ representing vertex set and edge set of the snapshot $\mathcal{G}_p$. Each snapshot $\mathcal{G}_p$ represents the scenario within a duration of $\tau$ time steps, with $
T_f=\tau * P$. Due to the heterogeneity in scenario elements and interactions, we use superscripts to distinguish different types of nodes and edges, \emph{i.e.,} $\mathcal{V}_p = \bigcup_{z } \mathcal{V}_p^z$ and $\mathcal{E}_p = \bigcup_{r} \mathcal{E}_p^r$. Here, $z \in \{0,1\} $ indexes node types as agent and lane, and $r \in \{0,1\}$ distinguishes edge types,  covering agent-agent edges and agent-lane edges. Correspondingly,  $A_p$ denotes the adjacency matrix of $\mathcal{G}_p$ and 
$X_p = \left[\bm{x_{1,p}}, \bm{x_{2,p}}, \dots, \bm{x_{n, p}} \right]^T$ indicates the feature matrix of nodes, where $\bm{x_{i,p}}$ is the feature of node $v_i$ in snapshot $\mathcal{G}_p$ and  $n=|\mathcal{V}_p|$.

\subsection{Scenario Encoding}
\label{sec:enc}
We employ static heterogeneous graphs to model observed scenarios. The input information of each scenario consists of the historical states of all the agents in the scenario and road networks. The historical state $\bm{s_{i,t}}$ of agent $i$ at timestep $t$  consists of its position, velocity, yaw, shape, agent type, \emph{etc.}. The road networks are represented with polylines, composed of sampled centerline points of lane segments. To facilitate information propagation among multi-agents, we adopt a scenario-centric coordinate system. All scenario elements are represented with the global coordinate system.

Inspired by Liang \emph{et al.} \citep{liang2020learning} in their work on scenario representation, the scenario encoder consists of an agent Transformer, a road GCN, and a FusionGCN. The agent Transformer is adopted to encode the historical motion features of each agent. Specifically, the state of each agent is first embedded with a Multi-Layer Perceptron (MLP). Then, the embedding is processed with a stack of time-wise attention layers to encode the spatio-temporal information of historical states. We take the output features of the last observation timestep $t=0$ as the motion feature of each agent.

To encode the road networks, we first construct a road graph in accordance with the topology of the roads. In particular, nodes represent centerline points; two nodes are linked with an edge if their corresponding lane segments are connected. We employ a multi-scale graph convolutional network, termed road GCN, to learn to represent the road graph. The road GCN consists of graph Transformer layers, graph pooling layers, and graph unpooling layers. It first learns to encode local information of road structures with graph transformer layers and produce point-wise features. Then, we obtain the polyline-wise features through graph pooling and aggregate information from neighboring polylines through the subsequent graph transformer layers. Finally, polyline-wise features are mapped back to point-wise features through graph unpooling, fused with intermediate point-wise features to produce the embeddings of each node, \emph{i.e.,} features of each lane segment.

We further employ a heterogeneous graph, following HeteroGCN \citep{gao2023dynamic} but a static one, to handle the diverse interactions in the scenario, like agent-agent, agent-lane, and lane-lane interactions. It is constructed based on the last observation and composed of two types of nodes, including agent nodes and lane nodes, as well as various types of edges connecting them. Notably, agent-lane edges are constructed with depth-first-search (DFS) to permit agents to explore potential driving areas in the future. With motion features of agents and embeddings of lanes as input features $H_0$ of corresponding agent nodes and lane nodes, the heterogeneous convolution operator defined in HeteroGCN is employed to capture diverse interactions in the observed scenarios.
\begin{equation}\label{eq:enc}
H_l = HeteroGCN(H_{l-1}, A),
\end{equation}
with $H_l$ indicating hidden features of the $l$-th graph convolutional layer and $A$ representing the adjacency matrix of the heterogeneous graph. The hidden feature produced by the last layer of the encoder is taken as the embedded features of each agent, represented as $\bm{h}_i$,  and $H = [\bm{h}_1, \bm{h}_2, \dots, \bm{h}_n]^T$ for all of the agents.

\subsection{Joint Motion Forecasting with Progressive Multiscale Decoding}

We model each evolving future scenario as a dynamic heterogeneous graph and exploit dynamic graph representation learning to facilitate the joint prediction of multi-agent future motion. Through a multi-scale prediction strategy, the dynamic graph adaptively improves the modeling of evolving context information and enhances the accuracy and consistency of joint prediction, as illustrated in Fig.~\ref{f:1}. 

\subsubsection{Progressive modeling of future scenarios}
As the agent navigates, the interactions with neighboring vehicles, pedestrians, and roads undergo dynamic variations. We utilize discrete-time dynamic heterogeneous graphs to characterize the evolving and diverse interactions within the scenario. Here, nodes correspond to distinct scenario elements, like agents and lane segments, while different types of edges represent various interactions, including agent-agent and agent-lane interactions. These interactions capture the mutual influence among agents (\emph{e.g.,} vehicles and pedestrians) together with the constraining and guiding effects imposed by the road structure. The attribute information and connectivity of nodes are subject to continuous modifications as the dynamic scenario unfolds.  However, future states of multi-agents are inherently unknown and characterized by uncertainty. To tackle this challenge, we employ \textbf{a progressive scenario modeling approach}, where the structure of the dynamic graph at the next timestep is constructed based on the previously decoded joint future states of the multi-agents.

Specifically, a future scenario is represented as a discrete-time dynamic graph $\mathcal{G}$, where its snapshots $\mathcal{G}_p= \{\mathcal{V}_p, \mathcal{E}_p\}_{p=1}^P$ sequentially model the future scenario information at each $\tau$ time steps, with the prediction horizon $T_f = \tau * P$. In particular, interactions in a scenario during the timesteps
$t \in  [\tau*(p-1)+1,  \tau*p$] are encoded with the topology of $\mathcal{G}_p$, $p=1, 2,\dots,P$. Specifically, we consider two kinds of nodes, including agents and lane segments, represented with $v^0_{p,i} \in \mathcal{V}^0$  and $v^1_{p,j} \in \mathcal{V}^1$, respectively. The agent2agent and lane2agent interactions are modeled with edge $e_{p}^0(v_{p,i}^0,v_{p,j}^0) \in \mathcal{E}^0_p$ and edge $e_{p}^1(v_{p,m}^1,v_{p,n}^0) \  \in \mathcal{E}^1_p$, respectively.

We construct the next snapshot $\mathcal{G}_p$ of the dynamic graph $\mathcal{G}$, given the agent features from their last predicted state at time $t= \tau*(p-1)$, $i=1,2, \dots, n$, and lane features. Specifically, agent2agent edges are constructed to establish a fully connected graph among the interesting agents in each scenario. Furthermore, we consider the distance between each interesting agent and its surrounding lane segments to construct dynamic lane2agent edges. For accelerating dynamic edge connection, we constrain the candidate lane nodes for each agent node within its neighbors obtained through DFS in data preprocessing (Sec.~\ref{sec:enc}). In particular, for each discrete snapshot $\mathcal{G}_p$, we first calculate the $\ell_1$ distance based on the predicted position of each agent at last timestep $t=\tau *(p -1)$, \emph{i.e,} $\bm{c}_{i, p} = \hat{c}_{t}(v_i^0) \ \forall v_i^0 \in \mathcal{V}_p^0$, with $\hat{c}_t(\cdot)$ indicating the predicted locations of agent nodes at time $t$, and the center point of candidate lane segments $\bm{c}_{j, p}$
\begin{equation}
    d_p(v_i^0, v_j^1)= \|\bm{c}_{i, p}-\bm{c}_{j, p}\|_1, \quad p=1,2,\dots, P. \\
\end{equation}
Then, an edge $e_{p}^1(v_j^1,v_i^0)$ is constructed to link the lane node $v_j^1$ to the agent node $v_i^0$ if  $d_p(v_i^0, v_j^1) < \epsilon$, with $\epsilon$ as an hyperparameter.

As described above, we sequentially construct the snapshots of dynamic graphs based on the previous predictions, thereby forming the foundation for forecasting multi-agent joint motion in the subsequent time steps. This  progressive mechanism permits to capture the evolving dynamics in future scenarios.

\subsubsection{Multiscale Decoding}
To address potential error accumulation in the progressive prediction framework, we further introduce \textbf{a multi-scale decoding scheme}. This scheme performs motion predictions by alternating between graph updates and predicted state refinements. Specifically, the prediction contains three steps: coarse prediction based on dynamic graph information, snapshot update according to the coarse prediction, and prediction refinement.

\textbf{Coarse Prediction}. Jointly predict  specific locations of future trajectories at a coarse granularity in the next $\tau$ time steps, \emph{i.e.,} just predicting the midpoint and last positions of multi-agents at $t\in [\tau * (p-1)+1, \tau*p]$, based on the features from the constructed snapshot $\mathcal{G}_p$.

\textbf{Snapshot update}. Based on agents' future states at time $t=\tau*p$ rather than  $t=\tau*(p-1)$ used in snapshot construction, we update the attributes and connections of agent nodes in  graph snapshot $\mathcal{G}_p$ and obtain  $\mathcal{G}_p^{'}$.

\textbf{Joint Prediction}. Taking in the features from updated snapshot $\mathcal{G}_p^{'}$,  a joint header finally outputs the complete joint states of agents $\hat{\bm{y}}_{i,t}$, $i=1,2,\dots, n$, during the time interval  $t \in [ \tau*(p-1)+1,  \tau*p]$. Repeat such procedure until $t= T_f$.

\subsubsection{Network Architecture}
As illustrated in Fig.~\ref{f:1}, the decoder consists of a temporal module, dynamic graph construction and update modules, two heterogeneous graph convolution modules for the dual-stage dynamic graphs, and several headers responsible for decoding the joint future states of interacting agents within a scenario. We elaborate on the main components below. 

In particular, we adopt \textbf{a factorized strategy to exploit the spatio-temporal information} within dynamic graphs. Building upon the embedded features of agents $H$ extracted from the encoder,  we first decode the temporal features corresponding to the future motion of the agents. To promote smooth and coherent motion trajectories, we employ cross-time attention to improve on temporal consistency among the features of each agent within the prediction horizon. Furthermore, we tackle the spatial interactions at each future timestep with designed heterogeneous graph convolution modules. With the help of message propagation in heterogeneous graph convolution, features from different nodes, corresponding to agents and lanes, are continuously aggregated and transformed, resulting in spatially consistent future features of the multi-agents. Finally, taking in these spatially and temporally consistent features of agents, the prediction header outputs the joint future states of multiple agents. 

\textbf{Temporal module.} To improve on the temporal consistency of future motions, we employ cross-time attention to handle the embedded agent features from the encoder, $\bm{h}_i, i=1,2,\dots, n$. Specially, to facilitate handling future interactions between agents and surrounding road networks in the subsequent module, we introduce a collection of latent variables to represent agent features at different future timesteps. Specifically, each agent feature $\bm{h}_{i} \in \mathbb{R}^{d}$ is transformed to a 2-D tensor $\bm{x}_{i} = [\bm{x}_{i,1}, \dots, \bm{x}_{i,t}, \dots, \bm{x}_{i, T_f}]^T \in \mathbb{R}^{T_f\times d}$ through 1-D convolutional layers,  corresponding to latent features of agent $i$ for future $T_f$ timesteps. Furthermore,  we resort to cross-time attention to handle their respective latent variables $\bm{x}_{i, t}, \ t=1,2, \dots, T_f$ in order to capture the temporal dependency of the future motion of each agent $\bm{x}_{i}$. Positional embedding is further imposed on
\begin{gather}
    \bm{x}_{i,t} = \bm{x}_{i,t} + PE(\bm{x}_{i,t})\\
    \bm{x}_{i} = {\rm Attention}(\bm{x}_{i}). \label{eq:temp}
\end{gather}

\textbf{Heterogeneous graph convolution operator.} The designed graph convolution operator, rooted in message-passing schemes \citep{GilmerSRVD17,hamilton2017inductive}, aggregates information from agent nodes and lane nodes along distinctive edges. Specifically, given snapshot $\mathcal{G}_p$, for an arbitrary agent node $\bm{v}_{i,p}^0$, messages $m(\cdot)$  that are adaptively computed from pairwise node features $\bm{x}_{\_,p}$ and node coordinates $\bm{c}_{\_,p}$ are firstly propagated within its two kinds of neighborhoods, \emph{i.e.,} agent neighborhood $N^0_p(v_i^0)$ and lane neighborhood $N^1_p(v_i^0)$, and aggregated with max-pooling. Then, the aggregated features are respectively transformed with functions approximated by MLPs $f_a(\cdot)$ as well as $f_l(\cdot)$,  and are added to the self-transformed features $g(\bm{x}_{i, p})$ to update the features of node $\bm{v}_{i,p}^0$
\begin{align}\label{e:hgcn}
\bm{z}_{i, p} = \ & 
 g(\bm{x}_{i, p})  + f_a \large(\max_{v_j^0  \in N^0_p(v_i^0)} m(\bm{x}_{i, p}, \bm{c}_{i, p-1}; \bm{x}_{j, p},  \bm{c}_{j, p-1} )\large) \nonumber \\
&+  f_l \large( \max_{v_l^1 \in N^1_p(v_i^0)} m(\bm{x}_{i, p}, \bm{c}_{i, p-1}; \bm{x}_{l, p},  \bm{c}_{l, p-1} \large),
\end{align}
where $g(\cdot)$ denotes self-transformation implemented with an MLP. Similarly, the graph convolution operator in the second graph convolution module takes in the updated snapshot $\mathcal{G}_p^{'}$ and produces agent node features $\bm{z}_{i,p}^{'}$.

\textbf{Prediction headers.} All the prediction headers are designed as MLPs. Specifically, the coarse header predicts the midpoint and last positions of multi-agents in the next $\tau$ time steps, \emph{i.e.,} $t \in [ \tau*(p-1)+1,  \tau*p]$, 
based on the agent features $\bm{z}_{i, p}$ in  $\mathcal{G}_p$, $i=1,2,\dots, n$, from the first heterogeneous graph convolution module. The joint header, taking in the agent features $\bm{z}_{i,p}^{'}$ in the updated snapshot $\mathcal{G}_p^{'}$  from the second heterogeneous graph convolution module, produces the complete joint states of agents,  $\hat{\bm{y}}_{i,t}$ for $t \in [ \tau*(p-1)+1,  \tau*p]$ and  $i=1,2,\dots, n$. 

Furthermore, to facilitate the temporal module in effectively capturing temporal information of the future motion of agents, we introduce a marginal prediction header. It takes in the agent features from Eq.~\eqref{eq:temp} and makes $K$ marginal predictions of the future state of each agent.

\textbf{Multimodal prediction.} Finally, given the uncertainties in the intentions and motions of agents, we make $K$ predictions for the joint motion of the multi-agents. To induce the model to perform consistent prediction within each modality, we introduce learnable scenario embeddings, $\bm{m}_k$, $k=1, 2, \dots, K$, to represent latent features of each scenario modality. In the same scenario modality, the agent features obtained from the encoder are combined with the corresponding scenario embeddings. This integration permits the model to leverage the individual characteristics of each agent alongside the overall context provided by the scenario embedding. Consequently, the predictions of all agents within the same scenario modality undergo a biasing process.
Mathematically, agent features $\bm{h}_{i}$ from the encoder become
\begin{equation}
    \bm{h}_{i,k} = \bm{h}_i + \bm{m}_k, \qquad i=1, 2, \dots, n
\end{equation}
where the subscript $k$ indicates the variable belonging to different modes.

\subsection{Training}
The model is trained to minimize the smooth $\ell_1$-norm loss between the joint prediction of agents' future states $\bm{\hat{y}}_{i,t, k}$ and corresponding ground truth $\bm{y}_{i,t}$.  Following previous methods, this loss is computed based on the best one among $k=1,2, \dots, K$ predictions, in terms of the Joint Final Displacement Error (JFDE) metric. Mathematically,
\begin{equation}
     \mathcal{L}_{joint}= \frac{1}{|\mathcal{V}^0|T_f}  \sum_{v_i^0 \in \mathcal{V}^0} \sum_{t=1}^{T_f} l(\bm{\hat{y}}_{i,t, k^*}, \bm{y}_{i,t}),
\end{equation}
with
\begin{equation}
    k^* = {\arg \min}_k  \sum_{v_i^0 \in \mathcal{V}^0} \| \hat{c}_{T_f, k}(v_i^0) - c_{T_f}(v_i^0) \|_2,
\end{equation}
where $c_{T_f}(\cdot)$ and $\hat{c}_{T_f,k}(\cdot)$ respectively represent the ground truth locations of agent nodes and the predicted ones in the $k$-th modality at time $T_f$. Similarly, the loss of coarse prediction is calculated and denoted as $\mathcal{L}_{mid}$.

Furthermore, an auxiliary loss $\mathcal{L}_{marg}$  corresponding to the marginal prediction is calculated in a similar manner to the joint prediction loss, except that each agent is computed with their respective best marginal prediction
\begin{equation}
    k^* = {\arg \min}_k  \| \hat{c}_{T_f, k}^{(m)}(v_i^0) - c_{T_f}(v_i^0) \|_2,
\end{equation}
where the superscript $\cdot^{(m)}$ refers to the marginal prediction.

The encoder-decoder networks are optimized with the weighted sum of such losses through gradient descent
\begin{equation}\label{e:loss}
    \mathcal{L} = \mathcal{L}_{joint} +  \lambda_1 * \mathcal{L}_{mid} + \lambda_2 * \mathcal{L}_{marg},
\end{equation}
where $\lambda_1$ and $\lambda_2$ are hyper-parameters to reconcile the three objectives.

\section{Experiments}
In this section, we compare ProgD with a collection of state-of-the-art joint motion forecasting methods on two large-scale real-world benchmarks and further investigate its various components through a series of ablation studies.

\subsection{Experimental Settings}
\textbf{Datesets.} We evaluate methods on two widely adopted real-world multi-agent joint motion forecasting benchmarks, INTERACTION Multi-Agent Track \citep{zhan2019interaction} and Argoverse 2 Multi-World Forecasting \citep{wilson2023argoverse}. The INTERACTION v1.2 dataset consists of highly interactive scenarios, including roundabouts, ramp merging, double lane change, and unsignalized intersections, collected from three continents (North America, Asia and Europe). Each scenario lasts for 4 seconds, with a 1-second observation followed by a 3-second prediction horizon (\emph{i.e.}, $T_o=1$ and $T_f=3$), sampled at 10 HZ. There are three types of agents in scenarios: car, pedestrian, and bicycle. Argoverse~2 \citep{wilson2023argoverse} is composed of 250,000 scenarios, sampled at 10HZ from six distinct cities in the United States. 
The official data split is provided with a ratio of 8:1:1 for the training, validation, and testing sets. Each scenario contains 5-second observations and a 6-second prediction horizon, \emph{i.e.}, $T_o=5$ and $T_f=6$. A case typically consists of 5 types of dynamic
agents and 5 types of static agents. In both datasets, there is an ego agent and multiple closely interacting agents, referred to as interesting agents, that are labeled for joint prediction in each scenario. The locations, speed, and heading of agents are rendered as agents' states $\{\bm{s_t}\}$, and HD maps are provided.

\begin{table*}[t]
\caption{Results on the INTERACTION multi-agent leaderboard. }
\label{t:int}
\centering
\setlength{\tabcolsep}{3.5mm}
\begin{tabular}{l|ccccc|c}
\toprule
Method &minJADE & minJFDE & minJMR & crossCR & egoCR & \textbf{Consis-minJMR}\\
\midrule
Graphformer&0.4572&1.4683&0.1728&0.0143&0.3450&0.4211\\
Mtz&0.3778&1.1649&0.2589&0.0773&0.0061&0.2721\\
ReCoG2 \citep{mo2020recog}&0.4668&1.1597&0.2377&0.0688&0.0106&0.2684\\
THOMAS \citep{gilles2022thomas}&0.4164&0.9679&0.1791&0.1277&0.0106&0.2516\\
HDGT \citep{JiaWCLLY23}&0.3030&0.9580&0.1938&0.1629&0.0049&0.2356\\
DenseTNT \citep{gu2021densetnt}&0.4195&1.1288&0.2240&\textbf{0.0000}&0.0136&0.2240\\
AutoBot \citep{GirgisGCWDKHP22}&0.3123&1.0148&0.1933&0.0430&0.0098&0.2068\\
HGT Joint&0.3070&1.0559&0.1864&0.0164&0.0053&0.1896\\
Traj-MAE \citep{ChenWSLHGCH23}&0.3066&0.9660&0.1831&0.0209&0.0064&0.1880\\
FJMP \citep{rowe2023fjmp}&\textbf{0.2752}&0.9218&0.1853&0.0052&0.0068&0.1866\\
MS-Net \citep{TangSHSG24}&0.3003&0.9619& 0.1832&-&-&-\\
EEGT~\citep{zhang2024edge}&0.34&1.08&0.2562&-&0.0044&-\\
\midrule
Proposed&0.3323&\textbf{0.8620}&\textbf{0.1538}&0.0094&\textbf{0.0011}&\textbf{0.1575}\\
\bottomrule
\end{tabular}
\end{table*}

\textbf{Metrics.} In accordance with the settings of both benchmarks, we make $K=6$ joint predictions, each joint prediction referred to as a modality, for interesting agents in each scenario and evaluate them with the following metrics. In the INTERACTION multi-agent benchmark, Minimum Joint Average Displacement Error (\textbf{minJADE}), Minimum Joint Final Displacement Error (\textbf{minJFDE}),  Minimum Joint Miss Rate (\textbf{minJMR}), Cross Collision Rate (\textbf{crossCR}), Ego Collision Rate (\textbf{egoCR}) and Consistent Minimum Joint Miss Rate (\textbf{Consis-minJMR}) are reported. Specifically, \textbf{JADE} represents the averaged $\ell_2$-norm distance between a modality and the ground truth over all the time steps and interesting agents, and \textbf{minJADE} denotes the minimum value over $K$ modalities. Similarly, \textbf{minJFDE} only considers the last predicted timesteps. For each scenario, a miss refers to a situation where the lateral or longitudinal final displacement error of an agent exceeds the corresponding threshold, and \textbf{JMR} as well as \textbf{minJMR} represent the ratio of miss for each modality and the minimum one among different modalities, respectively. \textbf{crossCR} represents the averaged collision frequency among the predictions over different modalities, while \textbf{egoCR} considers the collision between the ground truth of ego and the predictions of other agents. \textit{\textbf{Consis-minJMR}} as the ranking metric is similar to \textbf{minJMR}, except that the modality with a cross collision is taken as a miss. In Argoverse~2 motion forecasting benchmark, in addition to \textbf{minAFDE} and \textbf{minJFDE}, Actor Miss Rate (\textbf{actorMR}) and Actor Collision Rate (\textbf{actorCR}) are adopted as the miss ratio and the collision ratio of the best modality in terms of \textbf{minJFDE}, respectively. 
The ranking metric, \textit{Average Brier Minimum Final Displacement Error (\textbf{B-minJFDE})}, further considers the probability of the optimal modality. 

\textbf{Network architectures.} In the encoder, in consideration of the different observation windows, the Transformers for encoding the historical information of agents consist of $2$ and $4$ layers, each with $8$ attention heads, for the INTERACTION and Argoverse 2 datasets, respectively. For both datasets, the road GCN is composed of $4$ graph Transformer layers to encode road networks. $3$ heterogeneous graph convolution layers are employed to further process the spatio-temporal interactions in observed scenarios based on the embeddings of agents and lane segments from the Transformer and road GCN.

In the decoder, the temporal module consists of a 1-D convolution layer with kernel size $1$ and a 1-layer Transformer with $8$-head attention for both datasets. Both graph convolution modules contain $T_f/\tau$ proposed heterogeneous graph convolution layers, with $T_f$ as the prediction horizon and $\tau$ denoting the timeslot modeled by each snapshot of dynamic graphs. The marginal prediction and coarse prediction headers are designed as two-layer MLPs, and the joint prediction header is a three-layer MLP.

\textbf{Implementation details.} In the experiments, we utilize a scenario-centric coordinate system to facilitate information propagation among agents, where all agents within the scenario share a common coordinate system. Specifically, in the Argoverse 2 dataset,  we take the location of the ego agent at $t=0$ as the origin and align the current direction of the agent as the positive $x$-axis. In the INTERACTION dataset, a scenario is centered on the agent that is closest to the centroid of the positions of all the agents at $t=0$ and rotated accordingly to make its current direction as the positive $x$-axis. All the data including road networks and agent states are normalized accordingly. Regarding dynamic graph construction,  the time interval $\tau$, typically set as $1$ second, is studied in the ablation experiments, and the distance threshold $\epsilon$ for agent-agent edges is 15 meters. The hidden features of all the networks are set as $256$ for both datasets. Hyperparameters in the objective function Eq.~\eqref{e:loss} are configured as $\lambda_1 = \lambda_2 = 1.0$. We implement the proposed framework in Pytorch \citep{paszke2017automatic} and train it with the Adam \citep{kingma2014adam} optimizer for $90$ epochs on GeForce RTX 4090 GPUs. The batch size is $48$ and $32$ for the INTERACTION and Argoverse 2 datasets, respectively. 

\begin{table*}[!h]
\caption{Results on the Argovese~2 multi-world forecasting benchmark. }
\label{t:av2}
\centering
\setlength{\tabcolsep}{2mm}
\begin{tabular}{l|cc|cccc|c}
\toprule
&\multicolumn{2}{c|}{$K=1$}  &\multicolumn{5}{c}{$K=6$}\\
Method &minJADE & minJFDE & minJADE & minJFDE & actorMR & actorCR & \textbf{B-minJFDE}\\
\midrule
Lanegcn\_all&2.35&5.41&1.49&3.24&0.37&0.07&3.90\\
HiVT \citep{zhou2022hivt} &1.47&3.91&0.88&2.20&0.26&0.02&2.85\\
FJMP \citep{rowe2023fjmp} &1.52 &4.00&0.81&1.89&0.23&\textbf{0.01}&2.59\\
FFINet \citep{kang2024ffinet}&1.24&3.18&0.77&1.77&0.24&0.02&2.44\\
HistoryInfo&1.44&3.62&0.84&1.86&0.25&0.02&2.40\\
DGFNet \citep{xin2025multi}&-&-&0.73&1.68&0.20 &- &2.37\\
Forecast-MAE \citep{Cheng_2023_ICCV}&1.30&3.33&0.69&1.55&0.19&\textbf{0.01}&2.24 \\
SGPred&\textbf{1.10}&\textbf{2.77}&0.70&1.57&0.22&0.02&2.23\\
MIND~\citep{li2024multi}&-&-&0.70&1.62&0.20&\textbf{0.01}&-\\
\midrule
Proposed&1.24&3.12&\textbf{0.64}&\textbf{1.31}&\textbf{0.17}&\textbf{0.01}&\textbf{1.98}\\
\bottomrule
\end{tabular}
\end{table*}

\textbf{Baselines.} We compare the proposed 
ProgD with a series of state-of-the-art joint multi-agent motion prediction methods. First, graph neural network-based methods are taken into account. For instance, Lanegcn \citep{liang2020learning} and ReCoG2 \citep{mo2020recog} employ graph neural networks to encode observed scenarios, and decode the future trajectory with MLPs or RNNs. Based on the representation of observed scenarios from GNNs, DenseTNT \citep{gu2021densetnt} improves on decoding with offline optimization-based technique; THOMAS \citep{gilles2021gohome} recombines candidate marginal predictions to obtain consistent joint predictions with attention schemes; FFINet \citep{kang2024ffinet} introduces a feedback module to consider the influence
of future interactions; FJMP \citep{rowe2023fjmp} further exploits graph convolutional networks to perform marginal and conditional inference but is still limited to using static graphs to handle future interactions.  Furthermore, various Transformer-based methods are considered. For instance, hierarchical vector Transformers and heterogeneous graph Transformers are respectively explored in HiVT \citep{zhou2022hivt} and HDGT \citep{JiaWCLLY23} to handle the diverse interactions in observed scenarios. EEGT~\citep{zhang2024edge} further designs an edge-enriched graph transformer to handle the complex interactions.  AutoBots \citep{GirgisGCWDKHP22} implicitly handles future interactions through alternative attention layers across time and agents in the decoder. Similarly, Forecast-MAE \citep{Cheng_2023_ICCV} adopts a Transformer-based autoencoder and is trained in an unsupervised way. Built upon AutoBots, Traj-MAE \citep{ChenWSLHGCH23} resorts to masking strategies for consistent prediction, and MS-Net~\citep{TangSHSG24} exploits an evolutionary training process to enhance generalizability. Additionally, DGFNet~\citep{xin2025multi} considers  prediction difficulty among agents with a difficulty-guided decoder. MIND~\citep{li2024multi} further combines prediction with planning tasks.

\subsection{Experimental Results and Analysis}
\begin{figure}[!t]
\centering
 \subfigure[FJMP.]{ 
\centering
  \includegraphics[width=0.95\linewidth]{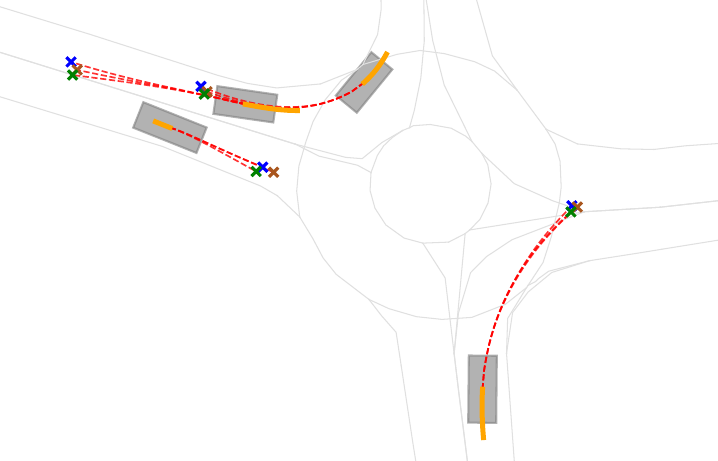}}
\vskip 0.2in
\centering
\subfigure[ProgD.]{ 
\centering
  \includegraphics[width=0.95\linewidth]{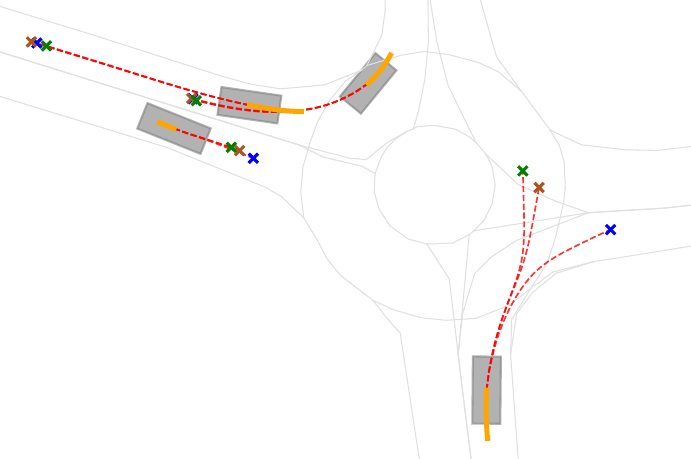}} 
 \caption{Visual comparison of joint prediction results on the testing set of INTERACTION. The pose of each agent at the last observation time is denoted by a gray rectangle.  Historical trajectories are depicted in orange,  and predicted trajectories are red dashed lines. The endpoints of the predicted trajectories are marked with `x' symbols, with each color indicating a different modality. Predictions for various agents within the same modality share the same color. For clarity, we present three modalities, \emph{i.e.,} three joint predictions.
 } \label{fig.vis.int}
\vskip -0.1in
\end{figure}

As presented in Table~\ref{t:int}, the proposed ProgD outperforms a collection of state-of-the-art methods \citep{rowe2023fjmp,ChenWSLHGCH23,TangSHSG24,zhang2024edge} on the INTERACTION multi-agent leaderboard in terms of the Consistent-minJointMR ranking metric, \textbf{ranking the $1^{st}$} from May 22, 2024 to March 14, 2025. As evaluated in terms of minJFDE and minJMR, ProgD achieves the best performance in accuracy. It reduces the minJFDE from 0.9218 to 0.8620 and decreases the minJMR from 0.1728 to 0.1538. This improvement is probably attributed to multiscale decoding, which prioritizes keyframe prediction (e.g., final positions within each time interval) before refining the complete trajectory. This strategy, while effective, may result in a slight increase in minJADE values. 
Furthermore, ProgD demonstrates competitive or superior performance in terms of consistency metrics, especially reducing the ego collision rate egoCR from 0.0044 to 0.0011. As shown in Fig.~\ref{fig.vis.int},  trajectories predicted by  ProgD demonstrate superior adherence to the road network structure, while also exhibiting consistency among highly interactive agents. For instance, Fig.~\ref{fig.vis.int} (a) illustrates that the baseline model erroneously predicts the bottom vehicle to be traveling in the wrong lane, whereas ProgD accurately predicts the vehicle to be in the correct lanes, as shown in  Fig.~\ref{fig.vis.int} (b).

\begin{table}[t]
\caption{Inference time per testing scenario on the INTERACTION dataset.}
\label{t:time}
\centering
\setlength{\tabcolsep}{4mm}
\begin{tabular}{l|c}
\toprule
Method& time (second)\\
\midrule
FJMP~\citep{rowe2023fjmp}&0.037 \\
\midrule
ProgD (static graph)&0.024\\
ProgD&0.032 \\
\bottomrule
\end{tabular}
\end{table}

\begin{figure*}[t]
\hskip 0.1in
\subfigure[ProgD.]{ 
\centering
\includegraphics[width=0.4\linewidth]{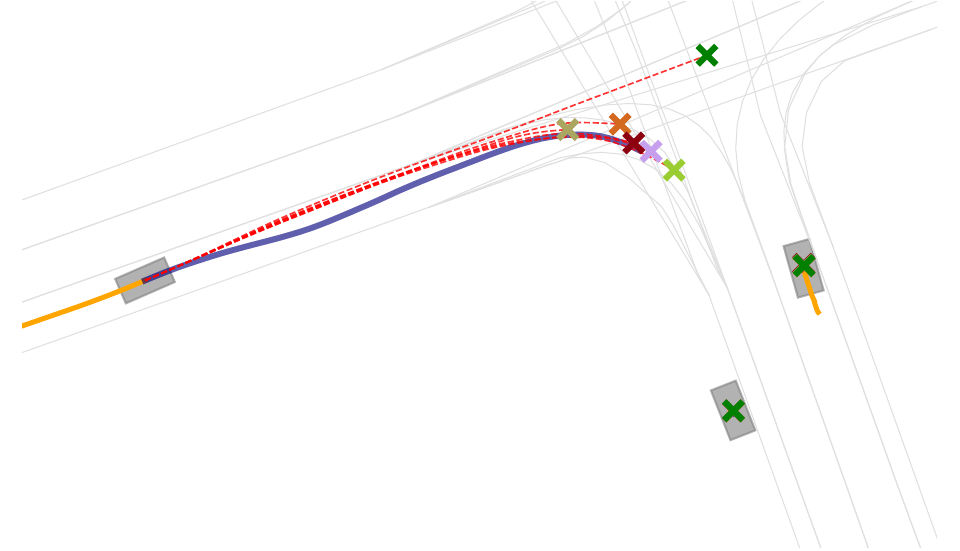}
}
\hskip 0.8in
 \subfigure[FJMP.]{ 
\centering
\includegraphics[width=0.4\linewidth]{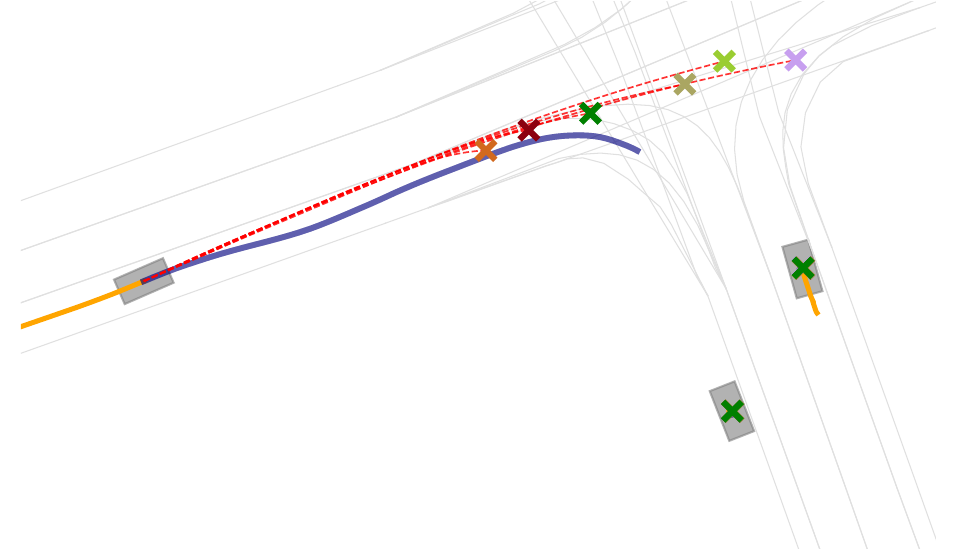}}\\
\hskip 0.1in
 \subfigure[ProgD.]{ 
\centering
\includegraphics[width=0.4\linewidth]{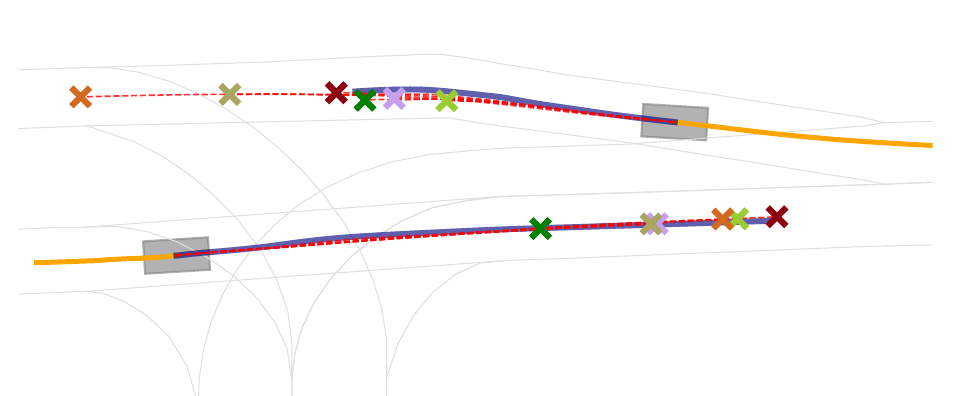}}
  \hskip 0.8in
 \subfigure[FJMP.]{ 
\centering
\includegraphics[width=0.4\linewidth]{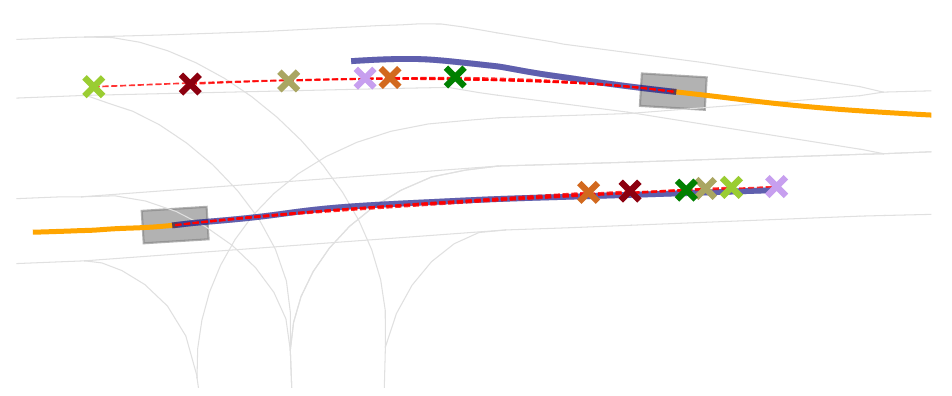}}
 \caption{Visual comparison of joint prediction results on the validation set of Argoverse~2.  The pose of each agent at the last observation time is denoted by a gray rectangle. Historical trajectories are depicted in orange, ground truth future paths in navy blue, and predicted trajectories are red dashed lines. The endpoints of the predicted trajectories are marked with `x' symbols, with each color indicating a different modality. Predictions for various agents within the same modality share the same color.
} \label{fig.vis.av2}
\vskip -0.05in
\end{figure*}

On the Argoverse 2 multi-world forecasting track, ProgD as a single model still achieves competitive performance, as shown in Table~\ref{t:av2}. It enhances the accuracy of multimodal joint prediction and improves probability estimation of the optimal modality, resulting in a reduction of the ranking metric average brier minimum final displacement error from 2.23 to 1.98. Meanwhile, each modality exhibits strong internal consistency, leading to a minimal actor collision rate 0.01. Fig.~\ref{fig.vis.av2} (a) and (b) demonstrate ProgD's superior accuracy in predicting future motion intentions and trajectories compared to the baseline model. Furthermore, the predictions from ProgD are more consistent with the road network structure compared to the baseline, as illustrated by the top vehicles in Fig.~\ref{fig.vis.av2} (c) and (d) better adhering to the lane center.

We further compare the inference time on testing set of the INTERSECTION dataset. Specifically, we measure the average inference time per scenario  utilizing a consumer-grade GPU (GTX 2080), with the official public code of the baselines. As shown in Table~\ref{t:time}, ProgD performs competitively with the static-graph baseline FJMP. Furthermore, we evaluate a variant of ProgD that employs static heterogeneous graphs to model future scenarios, excluding  progressive dynamic graph construction and multiscale decoding. This variant reduces the inference time from 0.032 seconds to 0.024 seconds, indicating that the additional time cost for progressive dynamic graph construction is acceptable.

\subsection{Ablation Studies}
In this section, we conduct ablation studies to assess the impact of various designs on joint prediction performance. Since the INTERACTION dataset does not provide a multi-agent version of the validation set, we conduct the ablation experiments on the Argoverse 2 validation dataset.

\begin{table*}[t]
\caption{Ablation studies on proposed modules on the validation set of Argovese~2 multi-world forecasting task. }
\label{t:abl}
\centering
\setlength{\tabcolsep}{1.4mm}
\begin{tabular}{c|ccccc|cc|ccc|c}
\toprule
\multirow{2}{*}{ID}&\multirow{2}{*}{Temporal}&G-Stat.&\multicolumn{2}{c}{G-Dyn.}&\multirow{2}{*}{Multiscale}& \multicolumn{2}{c|}{$K=1$}  &\multicolumn{4}{c}{$K=6$}\\
&&AA+AL&AA&AL&&minJADE&minJFDE&minJADE & minJFDE & actorMR & \textbf{B-minJFDE}\\
\midrule
1&$\checkmark$&&$\checkmark$&$\checkmark$&$\checkmark$&1.187&2.957&0.632&1.297&0.183&1.941\\
\midrule
2&&&&&&1.148&2.794&1.058&2.529&0.360&3.142\\
3&$\checkmark$&&&&&1.230&4.407&1.095&4.257&0.512&4.366\\
4&&&$\checkmark$&$\checkmark$&&1.232&3.066&0.654&1.339&0.191&1.983\\
5&$\checkmark$&&$\checkmark$&$\checkmark$&&1.183&2.950&0.639&1.317&0.188&1.958\\
\midrule
6&$\checkmark$&$\checkmark$&&&&1.176&2.930&0.653&1.348&0.192&1.991\\
7&$\checkmark$&&&$\checkmark$&&1.177&2.936&0.646&1.341&0.191&1.983\\
8&$\checkmark$&&$\checkmark$&&&1.173&2.930&0.650&1.352&0.193&1.995\\
\bottomrule
\end{tabular}
\vspace{0.2in}
\caption{Ablation studies on the time interval per snapshot in the dynamic graph. }
\label{t:tau}
\centering
\setlength{\tabcolsep}{5.5mm}
\begin{tabular}{c|cc|ccc|c}
\toprule
\multirow{2}{*}{}& \multicolumn{2}{c|}{$K=1$}  &\multicolumn{4}{c}{$K=6$}\\
&minJADE & minJFDE & minJADE & minJFDE & actorMR & B-minJFDE\\
\midrule
$\tau=3s$&1.207&3.017&0.638&1.315&0.187&1.964  \\
$\tau=2s$&1.188&2.962&0.632&1.300&0.183&1.945 \\
$\tau=1s$&1.187&2.957&0.632&1.297&0.183&1.941\\
$\tau=0.5s$&1.186&2.948&0.637&1.301&0.183&1.944 \\
\bottomrule
\end{tabular}
\vspace{0.2in}
\caption{Ablation studies on the auxiliary marginal prediction. }
\label{t:aux}
\centering
\setlength{\tabcolsep}{6mm}
\begin{tabular}{c|cc|ccc|c}
\toprule
\multirow{2}{*}{$\lambda_2$}& \multicolumn{2}{c|}{$K=1$}  &\multicolumn{4}{c}{$K=6$}\\
&minJADE & minJFDE & minJADE & minJFDE & actorMR & B-minJFDE\\
\midrule
$0$&1.261&3.135&0.653&1.342&0.193&1.984 \\
$0.2$&1.233&3.060&0.647&1.327&0.192&1.969 \\
$1.0$&1.187&2.957&0.632&1.297&0.183&1.941\\
$5.0$&1.153&2.870&0.632&1.298&0.182&1.943\\
\bottomrule
\end{tabular}
\end{table*}

The experimental results, as depicted in the second group of Table~\ref{t:abl}, demonstrate that a lightweight decoder (ID-2) experiences a significant decline in performance compared to the proposed method (ID-1). It utilizes an MLP to directly generate joint predictions based on the embeddings of historical states of agents and road networks from the encoder. Furthermore, relying solely on the temporal processing module (ID-3) does not yield performance improvements. Similarly, removing the temporal processing module and focusing exclusively on spatial interactions (ID-4) also leads to degraded performance. These findings emphasize the necessity of incorporating both temporal and spatial information in the decoding process. In addition, the architecture is further reinforced with multi-scale decoding, as shown by the comparison between ID-5 and the proposed method (ID-1).

We further study different strategies for future scenario modeling and representation learning. As shown in the last group of Table~\ref{t:abl}, using static heterogeneous graphs to model future scenarios (ID-6) results in a performance decline compared to the proposed heterogeneous dynamic graph-based approach. Furthermore, in dynamic graph-based scenario modeling, solely considering either agent-lane interactions  (ID-7) or agent-agent  interactions (ID-8) leads to performance degradation, especially without agent-lane interactions. These results suggest that dynamic modeling of the future social context, including road lanes and surrounding agents, benefits joint predictions.

In the study of progressive decoding,  as the time interval $\tau$ for encoding future scenarios of each snapshot in dynamic graphs is reduced from $3$ seconds to $1$ second, the performance of ProgD shows a gradual enhancement, as shown in Table~\ref{t:tau}. This improvement is attributed to the increased temporal resolution achieved by gradually reducing $\tau$, which enables finer-grained modeling of future scenarios and provides the model with more detailed and accurate dynamic information. However, it is crucial to consider the trade-off between temporal resolution and model complexity. Excessive reduction of $\tau$ increases model complexity and may potentially lead to overfitting.

Finally, as presented in Table~\ref{t:aux}, the auxiliary marginal prediction task benefits the joint prediction task. By directly predicting from temporal features in the marginal prediction task, the accompanying supervised signals enhance the temporal module in processing and exploitation of temporal dependency. The contribution of the auxiliary task is adjusted by the hyperparameter $\lambda_2$ in Eq.~\eqref{e:loss}, with the optimal performance at $\lambda_2=1.0$ as shown in  Table~\ref{t:aux}.

\section{Conclusion}
In this paper, we have introduced a progressive and multi-scale decoding strategy for accurate and consistent motion prediction of interacting agents in arbitrary numbers. It achieves excellent performance across multiple benchmark datasets, especially ranking the $1^{st}$ on the INTERACTION  multi-agent prediction benchmark. To explicitly capture evolving and diverse social interactions in future scenarios, we have designed a progressive modeling strategy of scenarios with dynamic heterogeneous graphs. This is complemented by an effective factorized architecture that exploits the spatio-temporal dependencies encoded in the graphs. Furthermore, the multi-scale procedure improves the capture and exploitation of complex interactions in future scenarios and enhances the accuracy of predictions. In the future, we will further explore the integration of joint multi-agent prediction and motion planning for autonomous vehicles.

\bibliographystyle{elsarticle-harv} 
\bibliography{ref}
\end{document}